\documentclass{article}
\usepackage{amsmath,epsfig,amsfonts,amssymb}
\usepackage{multirow,array}
\usepackage[pagebackref=true,breaklinks=true,colorlinks,bookmarks=false]{hyperref}

\usepackage[table,dvipsnames]{xcolor}
\usepackage[preprint]{spconf}
\copyrightnotice{\copyright\ IEEE 2019}
\toappear{To appear in {\it Proc.\ IGARSS 2019,
		July 28 - August 2, 2019, Yokohama, Japan}}
\title{SHIP INSTANCE SEGMENTATION FROM REMOTE SENSING IMAGES USING SEQUENCE LOCAL CONTEXT MODULE}
\name{Yingchao Feng\textsuperscript{1,2,3}, Wenhui Diao\textsuperscript{1,2}, Zhonghan Chang\textsuperscript{1,2,3}, Menglong Yan\textsuperscript{1,2}, Xian Sun\textsuperscript{*,1,2}, Xin Gao\textsuperscript{1,2}\thanks{This work is supported by National Natural Science Foundation of China under Grants 41701508.}\thanks{* Corresponding author: Xian Sun}}
\address{\textsuperscript{1}Institute of Electronics, Chinese Academy of Sciences, Beijing, China\\
\textsuperscript{2}Key Laboratory  of Network  Information System Technology (NIST), Institute of Electronics,\\Chinese  Academy of Sciences, Beijing, China\\
\textsuperscript{3}University of Chinese Academy of Sciences, Beijing, China}
\begin{document}
%\ninept
%
\maketitle
\begin{abstract}
The performance of object instance segmentation in remote sensing images has been greatly improved through the introduction of many landmark frameworks based on convolutional neural network. However, the object densely issue still affects the accuracy of such segmentation frameworks. Objects of the same class are easily confused, which is most likely due to the close docking between objects. We think context information is critical to address this issue. So, we propose a novel framework called SLCMASK-Net, in which a sequence local context module (SLC) is introduced to avoid confusion between objects of the same class. The SLC module applies a sequence of dilation convolution blocks to progressively learn multi-scale context information in the mask branch. Besides, we try to add SLC module to different locations in our framework and experiment with the effect of different parameter settings. Comparative experiments are conducted on remote sensing images acquired by QuickBird with a resolution of $0.5m-1m$ and the results show that the proposed method achieves state-of-the-art performance.
\end{abstract}
\section{Introduction}
\label{sec:intro}

Object instance segmentation aims to predict each pixel to different class label and distinguishes different objects of the same category. It can get more precise semantic information than object detection or semantic segmentation, so it has attracted much more attention in the remote sensing field. However, there are still many issues, such as confusion caused by the close docking between objects, making object instance segmentation still a challenging task.
\begin{figure}[htb]
	\centering
 	\centerline{\epsfig{figure=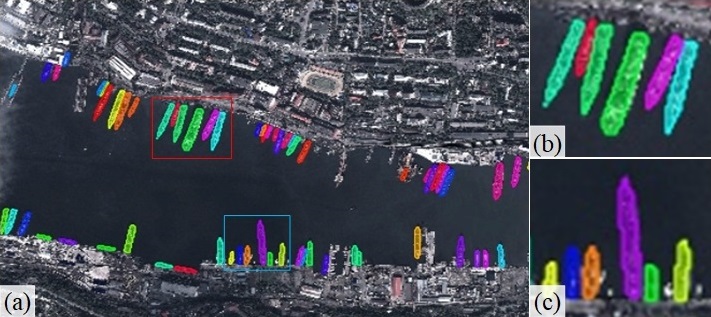,width=8.5cm,height=3.4cm}}
	\caption{An example is taken from the remote sensing image. (a) A typical image consisting of many dense ships. (b), (c) is the part of (a).}
	\label{fig:fig1}
\end{figure}

With the rapid development of Deep Convolutional Neural Networks (DCNN), the image features extracted by DCNN gradually replaced the features of the artificial design. Some powerful network structures, such as Mask R-CNN\cite{maskrcnn}, have proved the success of DCNN. But in our task, ship instance segmentation still exists an urgent problem to be solved. As shown in Fig. \ref{fig:fig1}, the remote sensing image with a resolution of $0.8m$ and a size of $16k\times16k$ have low resolution and large size compared with general natural scene image. The ships in this image only have a few hundred pixels or even dozens of pixels, and ships in the harbor are always densely docked. So, the small slices obtained by the bounding-box detection always contain the complete ship and other incomplete ships and easily classify the pixels of multiple ships into one class. This problem leads to confusion between different ships, greatly reducing the accuracy of ship instance segmentation.
\begin{figure*}
	\centering
	\centerline{\epsfig{figure=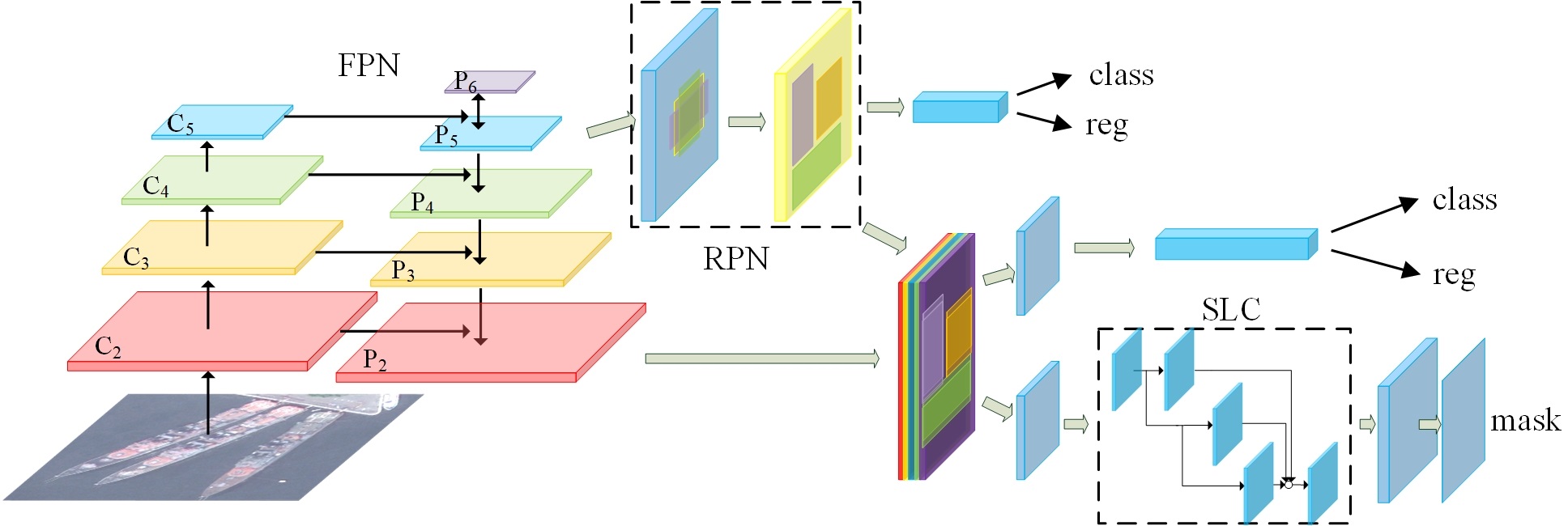,width=13.5cm,height=4.5cm}}
	\caption{Overview of our framework. The proposed sequence local context (SLC) module in mask branch to learn multi-scale context information.}
	\label{fig:fig2}
\end{figure*}

Context information is crucial for semantic segmentation to reduce ambiguous cases and obtain robust outputs. In remote sensing images, we can only see the top view of the ship. Therefore, the context information of the ship size, relative position and scene constraints can help to better segment different instances. More and more related works are focused on how to utilize context information. CCL\cite{ccl} fuses the features of two branches to get the context contrasted local feature. Inception\cite{inception} utilizes multi-channel convolution with different kernel sizes to combine the multi-scale information. ASPP\cite{aspp} employs different dilated convolution to extract different degrees of context information. However, due to the feature map size of the mask branch, the use of heavy structures or large kernel sizes and dilation rates is not suitable for our task. So, we present a novel instance segmentation framework to improve the mask-level AP in remote sensing  images.

To summarize, a context module called sequence local context (SLC) module is introduced to address the problem of ships confusion. The multi-layer structure consists of a sequence of dilation convolutional blocks, which progressively enhances the connection of context information at different scales and avoids the loss of information caused by dilation rate. Moreover, we experiment with how to better add the SLC module to our framework. After comparative and ablation experiments in different settings. Experiments show that our method obtains state-of-the-art performance.

\section{PROPOSED METHOD}
\label{sec:proposed method}

In remote sensing images, objects appear in arbitrary orientations and the distribution of objects is very uneven, for example, ships in the harbor are crowded, ships at sea are sparse. So, it is still a big challenge to get better instance segmentation performance of ships in remote sensing images. In this section, we leverage the context information to improve the performance of ship instance segmentation. The base framework is introduced in section \ref{ssec:base fram}, and the sequence local context (SLC) module is discussed in section \ref{ssec:SLC module}.

\subsection{Base Framework}
\label{ssec:base fram}

The overall framework of our instance segmentation network is shown in Fig. \ref{fig:fig2}. Mask R-CNN \cite{maskrcnn} gets the outstanding performance in many datasets, which based on Faster RCNN \cite{faster} and add a branch to predict mask. To combine the semantically strong features with semantically weak features, the FPN \cite{fpn} with ResNet-101 \cite{resnet} as the backbone is employed. And use RoIAlign layer to removes the error of harsh quantization. In this paper, the main structure of Mask R-CNN \cite{maskrcnn} is adopted as the backbone of our framework.

\subsection{Sequence Local Context Module}
\label{ssec:SLC module}

In our ship instance segmentation task, a big problem is ship density as shown in Fig. \ref{fig:fig1}, the small slices obtained by bounding-box detection always contain the complete ship and part of other ships, we want pixels belonging to incomplete ships to be background.

\begin{figure}[htb]
	\centering
	\centerline{\epsfig{figure=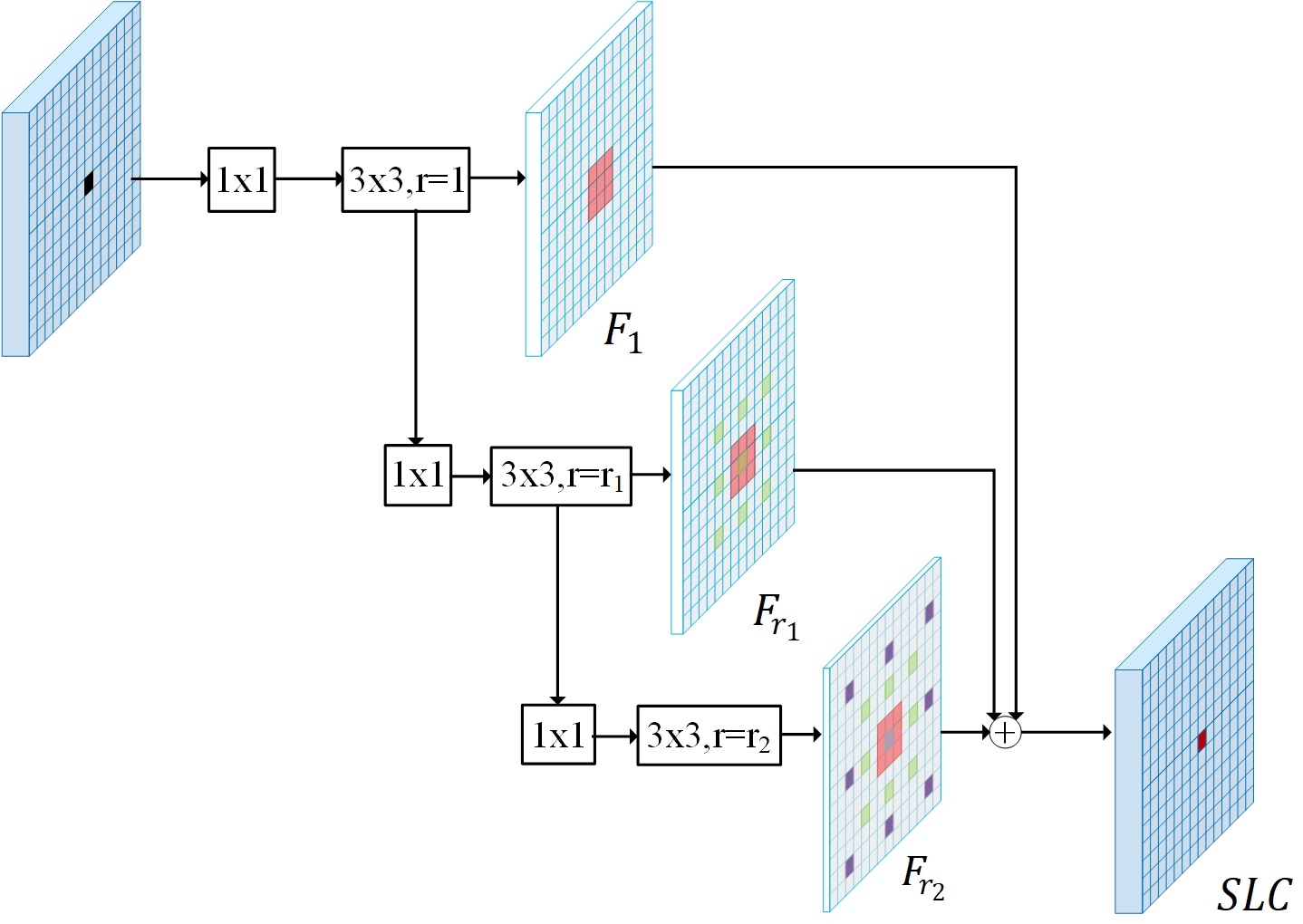,width=7.3cm,height=4.7cm}}
	\caption{Sequence Local Context (SLC) Module. We apply a sequence of dilation convolution blocks to combine multi-scale context information.}
	\label{fig:fig3}
\end{figure}

To address this problem, we propose a novel sequence local context module. Specifically, we utilize the dilated convolution with different dilation rate integrates multi-scale context features as shown in Fig. \ref{fig:fig3}, the SLC module consists of three layers, each with a 1$\times$1 convolution followed by a 3$\times$3 dilated convolution with different dilation rate to captures context information at different scales from the feature map. And then, we fuse the outputs of three layers by element-wise summation operation:
\begin{equation}
	SLC=F_1\left(\mathbb{F},\Theta_1\right)+F_{r_1}\left(\mathbb{F}_1,\theta_{r_1}\right)+F_{r_2}\left(\mathbb{F}_{r_1},\theta_{r_2}\right)
\end{equation}
where $\mathbb{F}$ is the input feature, $F_1$, $F_{r_1}$, $F_{r_2}$ are the function of convolution blocks with dilation rate 1, $r_1$,$r_2$, respectively. $\mathbb{F}_1$, $\mathbb{F}_{r_1}$ are the output feature of $F_1$, $F_{r_1}$. $\Theta_1$, $\Theta_{r_1}$ and $\Theta_{r_2}$ are respective parameters, and $SLC$ is the output feature. Through the sequence way, we can get discriminative-context to avoid misclassification the pixel of incomplete ships.

The receptive field size $R$ with dilated convolution can be calculated as follow:
\begin{equation}
	R=\left(r-1\right)\times\left(k-1\right)+k
\end{equation}
where $r$ is dilation rate, $k$ is kernel size, and if the convolution layer with receptive field size $R_1$ followed by convolution layer with receptive field size $R_2$, we can calculate the overall receptive field size $R_o$:
\begin{equation}
	R_0=R_1+R_2-1
\end{equation}

Using a large dilation rate directly would abandon lots of information and convolutions in the higher layer can employ features from the lower layers to enlarge receptive field. So, we select small dilation rate to avoid losing too much information and apply a sequence of convolution blocks to get large receptive field size, which progressively enhances the relationship between the local feature and the context information at different scales. In SLC module, the three layers have different receptive field sizes, which are $3$, $5+2\left(r_1-1\right)$ and $3+2\left(r_1+r_2\right)$ respectively. Based on our task and verified by experiments, we set the dilated rate $r_1=2$ and $r_2=3$.
\section{EXPERIMENTS AND RESULTS}
\label{sec:experiments}
\subsection{Dataset and Settings}
\label{ssec:dataset}

Our dataset is acquired by QuickBird with a resolution of $0.5m-1m$. The image size is $16k\times16k$ pixels. Centering on the position of each ship, the image was divided into $1k\times1k$ slices, and then the high overlap slices were removed using non-maximum suppression (NMS) with a threshold of 0.1, resulting in 2474 images and approximately 6500 objects. The ratio of the training set to test set is about 4:1. In addition, we utilize a random augmentation strategy. Specifically, we randomly change the brightness, contrast, color and sharpness of the image and then randomly rotate the image.

\begin{figure}[htb]
	\begin{minipage}[b]{.48\linewidth}
		\centering
		\centerline{\epsfig{figure=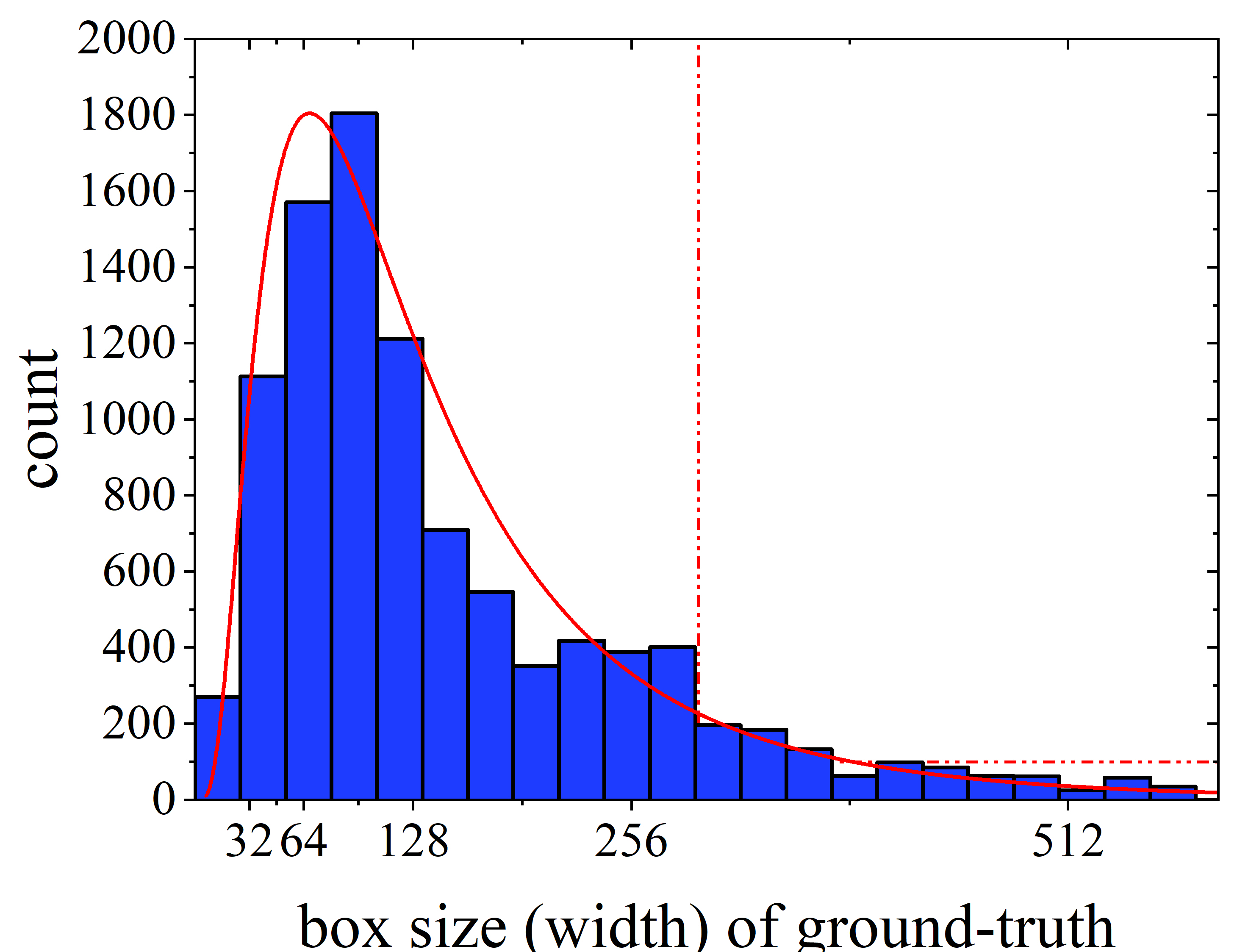,width=4.0cm,height=3.0cm}}
	\end{minipage}
	\begin{minipage}[b]{0.48\linewidth}
		\centering
		\centerline{\epsfig{figure=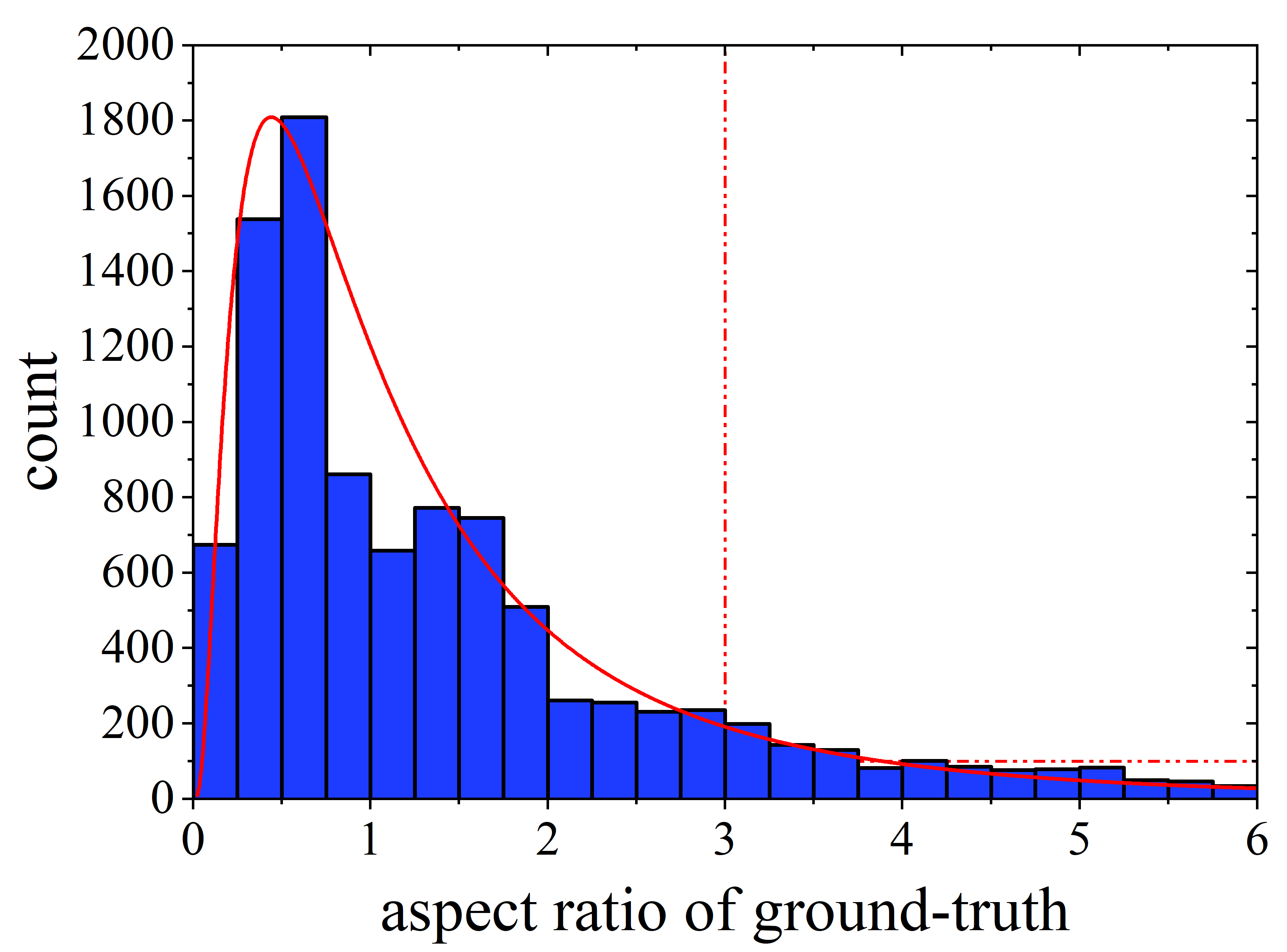,width=4.0cm,height=3.0cm}}
	\end{minipage}
	\caption{The distributions of the ship dataset: (a) the distribution of ship scales. (b) the distribution of aspect ratio.}
	\label{fig:fig4}
\end{figure}

The pixel sizes and aspect ratios of all ships in our dataset are shown in Fig. \ref{fig:fig4}. According to statistical results, we set our anchors from ${32}^2$ to ${512}^2$ by a multiple of 2 on different FPN \cite{fpn} layer, and add an extra anchor scale 0.707 to reduce the spacing of the anchor scales. In addition, we select [0.5, 1, 1.5] as the aspect ratios.

We re-implement Mask R-CNN \cite{maskrcnn} based on Keras, and use the pretraining model for MS COCO to initialize the network. During the training process. We trained the model for 25 epochs. The learning rate was set as 5e-4, Weight decay and momentum were 0.0001 and 0.9, respectively. After the RPN stage and NMS, we sampled 2000 ROIs in training and 1000 ROIs in inference for the second stage, then we select 200 proposals per image for training, where the ratio of positive to negative samples was 1:2, and 100 instances per image for predicting mask. Other parameters not mentioned follow the setting of Mask R-CNN \cite{maskrcnn}.
\begin{table}[htb]
	\centering
	\caption{Performance of SLC with different parameter settings, the augmentation and anchor setup strategies are used for all experiments.}
	\label{tab:tab1}
	\begin{tabular}{@{}lcc|cc@{}}
		Network&R(\%)&AP(\%)&R$^{bb}$(\%)&AP$^{bb}$(\%)\\
		\hline
		Mask R-CNN \cite{maskrcnn}&76.47&67.63&84.78&76.18\\
		+ inception v1 \cite{inception}&76.98&68.80&83.38&75.79\\
		+ dilated Conv&78.13&69.65&86.45&78.25\\
		+ CCL \cite{ccl}&81.71&73.36&85.17&77.27\\
		+ ASPP(2,4,6) \cite{aspp}&81.58&73.48&\textbf{86.57}&\textbf{79.86}\\
		+ SLC&\textbf{85.68}&\textbf{78.61}&86.32&79.65\\
	\end{tabular}
\end{table}
\subsection{Comparative Experiment}
\label{ssec:comparative}

We performed a series of experiments on ship datasets, and our method achieved state-of-the-art performance: 85.68\% for Recall and 78.61\% for AP. The experimental results of various methods are shown in Table \ref{tab:tab1}. The following will compare the different methods and analyze the results.

\begin{figure}[htb]
	\centering
	\centerline{\epsfig{figure=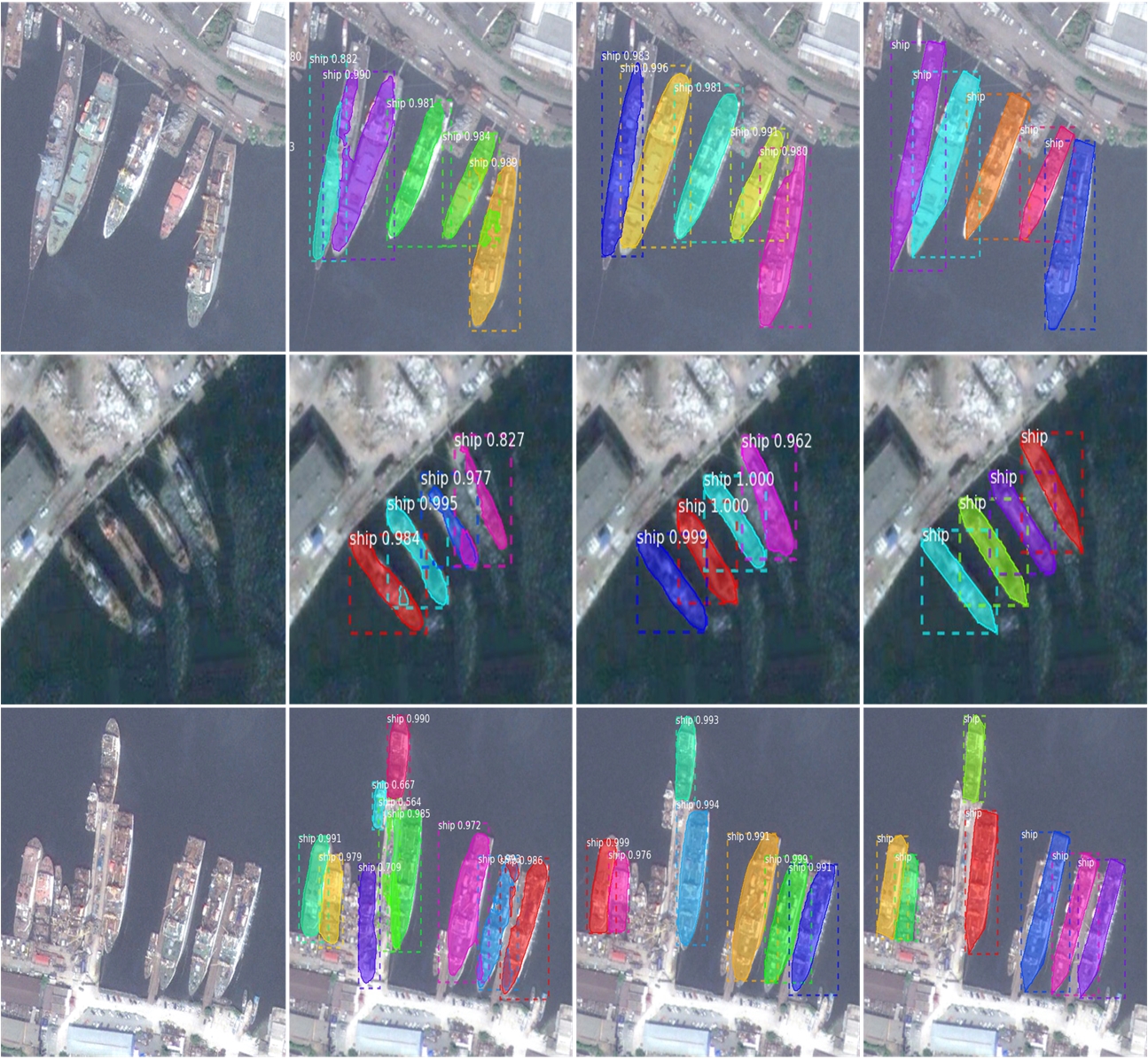,width=8.5cm,height=6.72cm}}
	\caption{The result of ship instance segmentation. For each image, we show result use Mask R-CNN \cite{maskrcnn} (middle left), SLC module (middle right) and the ground truth (right).}
	\label{fig:fig5}
\end{figure}

Due to the complexity of remote sensing images. Despite the use of data augmentation and modification of anchor and aspect ratio settings, the result we use Mask R-CNN \cite{maskrcnn} directly is not ideal. As shown in Fig. \ref{fig:fig5}, we find that the dense and sloping ships obtained by bounding-box detection always classify the pixels belonging to different ships into one category. We think this is because there is not enough context information.

Firstly, we add inception \cite{inception} structure on the mask branch, the improvement in performance proves our idea that context information can reduce ambiguous cases. Then, the dilated convolution enters our sight, we simply replaced the big convolution kernel with dilated convolution in inception \cite{inception} structure. The experiment results show our ideas that employ dilated convolution and combine multi-scale features can better extract context information in our task.

The CCL \cite{ccl} and ASPP \cite{aspp} are in line with our ideas. Based on our task, we modify the dilation rate of ASPP \cite{aspp} from (6,12,18) to (2,4,6), Table \ref{tab:tab1} shown that the results of both are similar and got a big boost. Then, based on the above experiment and the discussion in section \ref{ssec:SLC module}, we proposed the SLC module for our work. The experiment shows that our SLC module can extract the context information better and get state-of-the-art performance.

We also compare SLC module with others at the bounding-box object detection. As shown in Table \ref{tab:tab1}, context information doesn’t seem to have a significant impact on the performance of object detection relative to semantic segmentation. the best result only increases the margin of 3.0 points box-level AP over the Mask R-CNN. The performance of SLC drops slightly by 0.21\%. But our SLC module improves performance by 10.98\% on mask-level AP, which attains a small gap between its mask-level and box-level AP.
%\begin{table}
%	\centering
%	\caption{Performance of SLC module with different parameter settings.}
%	\label{tab:tab2}
%	\resizebox{240pt}{40pt}{
%		\begin{tabular}{l|cccc|cc}
%			Method&layers=2&layers=3&cls\&reg&$r_{1}$\&$r_{2}$&R(\%)&AP(\%)\\
%			\hline
%			\multirow{5}*{SLC} &\checkmark&&&2,3&78.64&70.23\\
%			&&\checkmark&&2,3&\textbf{85.68}&\textbf{78.61}\\
%			&&\checkmark&\checkmark&2,3&80.56&74.67\\
%			&&\checkmark&&2,4&78.26&70.38\\
%			&&\checkmark&&2,4&76.47&62.51\\
%	\end{tabular}}
%\end{table}

\subsection{Ablation Analysis}
\label{ssec:ablation}
In Table \ref{tab:tab2}, we experiment with the effect of different parameter settings on the SLC module. We first fix the dilation rates $r_1,r_2=2,3$, change the number of layers that are fused. Employing three layers is better than fusing the first and third layers. Then, we add SLC module to the classification and regression branch at the same time, we give up on this idea because of the performance degradation. Finally, we experiment with different dilation rates, as we expect that the greater the dilation rates, the worse the performance. On the one hand, the large dilation rates can’t learn the relationship of context information at different scales. On the other hand, a too large dilation rate will degenerate to $1\times1$ convolution and can’t capture the context information.
\section{CONCLUSION}
\label{sec:conclution}

In this paper, we focus on the confusion of objects in ship instance segmentation. To avoid the pixels of different ships dividing into one category, we verify that learning context information is essential to solve this problem and propose the sequence local context (SLC) module to progressively learn multi-scale context information. In our module, we learn the context information through applying the dilation convolution in sequence, and our module gets better performance in the remote sensing images. Experimental results show that the context information learned by SLC module is crucial to get the better performance. As future work, we will try to extend our method to verify its generalization in public datasets.
\begin{table}
	\centering
	\caption{Performance of SLC module with different parameter settings.}
	\label{tab:tab2}
	\begin{tabular}{@{}p{1.3cm}<{\centering}@{}|@{}p{1.4cm}<{\centering}@{}p{1.3cm}<{\centering}@{}p{1.3cm}<{\centering}@{}p{1.1cm}<{\centering}@{}|@{}p{1.1cm}<{\centering}@{}p{1cm}<{\centering}@{}}
		Method&layers=2&layers=3&cls\&reg&$r_{1}$\&$r_{2}$&R(\%)&AP(\%)\\
		\hline
		\multirow{5}*{SLC} &\checkmark&&&2,3&78.64&70.23\\
		&&\checkmark&&2,3&\textbf{85.68}&\textbf{78.61}\\
		&&\checkmark&\checkmark&2,3&80.56&74.67\\
		&&\checkmark&&2,4&78.26&70.38\\
		&&\checkmark&&2,4&76.47&62.51\\
	\end{tabular}
\end{table}

% Below is an example of how to insert images. Delete the ``\vspace'' line,
% uncomment the preceding line ``\centerline...'' and replace ``imageX.ps''
% with a suitable PostScript file name.
% -------------------------------------------------------------------------

%\vfill
%\pagebreak

\bibliographystyle{IEEEbib}
%\small
\bibliography{refs}
%}

\end{document}